\documentclass{article}

\usepackage[preprint]{neurips_2024}

\usepackage[utf8]{inputenc} 
\usepackage[T1]{fontenc}    
\usepackage{hyperref}       
\usepackage{url}            
\usepackage{graphicx}       
\usepackage{epstopdf}
\usepackage{amsfonts}       
\usepackage{amsmath}
\usepackage{mathtools}
\usepackage{array}
\usepackage{tabularx}
\usepackage{booktabs}       
\usepackage{multirow}
\usepackage{makecell}
\usepackage{wrapfig}
\usepackage[titletoc]{appendix}
\usepackage[linesnumbered,ruled,vlined]{algorithm2e}
\usepackage{xcolor}         
\usepackage{microtype}      
\usepackage[export]{adjustbox}
\usepackage{pifont}
\usepackage[subtle]{savetrees}
\usepackage{float}  

\definecolor{mygreen}{rgb}{0.0,0.541,0.059} 
\definecolor{mypurple}{rgb}{0.8667,0.2118,0.7059}
\definecolor{mypink}{rgb}{1, 0.851, 0.851}
\definecolor{myblue}{rgb}{0.0667,0.4431,0.8980}
\definecolor{myorange}{rgb}{0.9882,0.5804,0.1922}
\definecolor{lightyellow}{rgb}{1,1,0.88} 
\definecolor{lightred}{rgb}{1,0.88,0.88} 
\definecolor{lightblue}{rgb}{0.749,0.749,1} 
\definecolor{OliveGreen}{rgb}{0.0, 0.5, 0.0}

\newcommand{\mathbold}[1]{\ensuremath{\boldsymbol{\mathbf{#1}}}}

\newcommand{\g}{\,|\,}

\newcommand{\nestedmathbold}[1]{{\mathbold{#1}}}

\newcommand{\mbe}{\nestedmathbold{e}}

\newcommand{\mbx}{\nestedmathbold{x}}

\newcommand{\mbz}{\nestedmathbold{z}}

\newcommand{\mbD}{\nestedmathbold{D}}

\newcommand{\mbI}{\nestedmathbold{I}}

\newcommand{\mbphi}{\nestedmathbold{\phi}}

\newcommand{\mbtheta}{\nestedmathbold{\theta}}

\newcommand{\mbone}{\nestedmathbold{1}}

\DeclareRobustCommand{\KL}[2]{\ensuremath{\textsc{kl}\left[#1\;\|\;#2\right]}}

\newcommand{\cL}{\mathcal{L}}
\newcommand{\cN}{\mathcal{N}}

\newcommand{\E}{\mathbb{E}}

\title{Formatting Instructions For NeurIPS 2024}

\author{%
  Zhangkai Wu,  $^{\heartsuit,\clubsuit}$ \\
  \texttt{zhangkai.wu@mq.edu.au} \\
  \And
  Xuhui Fan$^{\clubsuit}$ \\
  \texttt{xuhui.fan@mq.edu.au} \\
  \AND
  Jin Li$^{\heartsuit}$ \\
  \texttt{cvjinli@outlook.com} \\
  \And
  Zhilin Zhao$^{\clubsuit}$ \\
  \texttt{zhilin.zhao@mq.edu.au} \\
  \And
  Hui Chen$^{\clubsuit}$ \\
  \texttt{hui.chen2@students.mq.edu.au} \\
  \And
  Longbing Cao$^{\clubsuit}$ \\
  \texttt{longbing.cao@mq.edu.au} \\
  $^{\heartsuit}$University of Technology Sydney\\
  $^{\clubsuit}$Macquire University
}

\title{ParamReL: Learning Parameter Space Representation  via Progressively Encoding Bayesian Flow Networks}

\begin{document}

\maketitle

\begin{abstract}

The recently proposed Bayesian Flow Networks~(BFNs) show great potential in modeling parameter spaces, offering a unified strategy for handling continuous, discretized, and discrete data. However, BFNs cannot learn high-level semantic representation from the parameter space since {common encoders, which encode data into one static representation, cannot capture semantic changes in parameters.} This motivates a new direction: learning semantic representations hidden in the parameter spaces to characterize mixed-typed noisy data. {Accordingly, we propose a representation learning framework named ParamReL, which operates in the parameter space to obtain parameter-wise latent semantics that exhibit progressive structures. Specifically, ParamReL proposes a \emph{self-}encoder to learn latent semantics directly from parameters, rather than from observations. The encoder is then integrated into BFNs, enabling representation learning with various formats of observations. Mutual information terms further promote the disentanglement of latent semantics and capture meaningful semantics simultaneously.} We illustrate {conditional generation and reconstruction} in ParamReL via expanding BFNs, and extensive {quantitative} experimental results demonstrate the {superior effectiveness} of ParamReL in learning parameter representation. We release the code at: \url{https://github.com/amasawa/ParamReL/tree/main}. 

\end{abstract}

\section{Introduction}
\label{sec:intro}

Representation learning~\cite{bengio2013representation}, which aims at learning low-dimensional latent semantics from high-dimensional observations, {offers an unsupervised approach to discovering high-level semantics in observations.} It has been widely applied in areas such as computer vision~\cite{CVPRLiCMZKK23,CVPRZhaoWL23,CVPRDongHLLQG23}, and data analytics~\cite{tonekaboni2022decoupling,oublal2024disentangling}. While most representation learning methods~\cite{kim2018disentangling,chen2018isolating,meo2023alpha} work on continuous-valued observations, different non-trivial methods are needed to discover semantics for the discretized ~\cite{van2017neural,razavi2019generating} and discrete data~\cite{austin2021structured,chen2022analog}. {Consequently, these individual efforts might face issues such as inconsistent discoveries within the data~\cite{zhou2024beta} or repeated modelling efforts~\cite{krishnan2018challenges,zhao2020variational}.} 

On the other hand, Bayesian Flow Networks~(BFNs)~\cite{graves2023bayesian,song2024unified,xue2024unifying} have been recently proposed as a promising deep generative model. By operating in the parameter space, BFNs design a multi-step mechanism to approximate the ground-truth parameters of observation sequentially. As a result, a uniform strategy may be adopted to deal with continuous, discretized, and discrete data while simultaneously maintaining fast sampling. Pilot studies of BFNs have shown promising results in modelling different data formats.

Leveraging BFNs, this paper introduces ParamReL, a novel parameter space representation learning framework that employs a unified strategy to extract meaningful high-level semantics from continuous, discretized, and discrete data. Specifically, a \emph{self-}encoder is designed to encode step-wise parameters into lower-dimensional latent semantics, capturing gradual semantic changes throughout the multi-step generation process. These latent semantics are then integrated into a neural network architecture to form the parameters for an output distribution that simulates observations. Furthermore, mutual information is introduced to enhance the disentanglement of latent semantics, promoting the capture of distinct and meaningful representations.

ParamReL is applied on benchmark datasets and verifies its effectiveness in obtaining meaningful high-level semantics for discrete and continuous-valued observations. Sampling and reverse-sampling procedures are developed here to complete conditional image reconstruction and generation tasks. In particular, our developed self-encoder discovers interesting progressive semantics along with the flow steps. That is, our ParamReL obtains meaningful and  clearer disentangled representations while maintaining high sample quality.

The main contributions of this work can be summarized as follows: (1) A parameter space representation learning framework introduces a uniform strategy for modelling continuous, discretized and discrete observations; (2) A \emph{self-}encoder encodes step-wise parameters into step-wise semantics to reveal a series of gradually changing latent semantics; (3) A mutual information term promotes latent semantics being disentangled and storing meaningful semantics simultaneously; (4) Sampling and reverse-sampling methods are developed, and generation and reconstruction tasks are completed in the parameter space. 
 
\section{Bayesian Flow Networks}
Bayesian Flow Networks~(BFNs)~\cite{graves2023bayesian,song2024unified,xue2024unifying} are a new class of latent variable models. {Similar to the reverse process of diffusion models \cite{ho2020denoising,song2020denoising}, BFNs target optimizing the neural network architecture on posterior parameters to simulate observations as they are gradually observed.} In particular, BFNs assume two types of distributions: a simple \emph{input distribution} $P_\mathrm{I}(\cdot)$ representing the initial belief about observations and an \emph{output distribution} $P_\mathrm{O}(\cdot)$ representing the simulating distribution for observations. The parameters of input distribution {are first updated} through a Bayesian inference scheme and then passed into a neural network to form the parameters of output distributions. The main objective of BFNs is to minimize the {Kullback-Leibler (KL) divergence between the ground-truth data distribution and the output distribution, ensuring that the output distribution closely approximates the ground-truth data distribution.}

\begin{figure}[b]
    \centering
    \includegraphics[width=0.9\columnwidth]{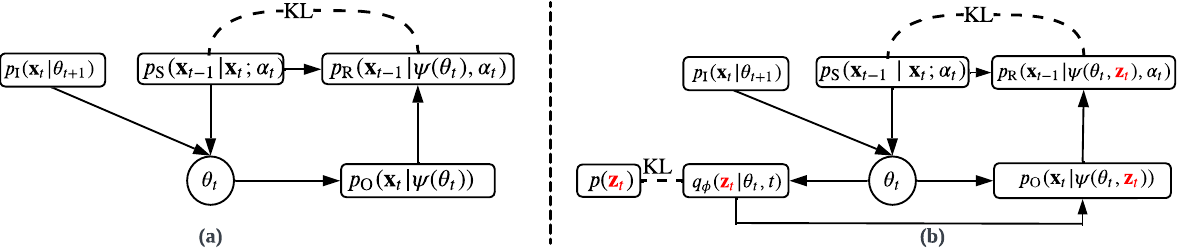}
    \caption{The relationships between distributions in BFNs~(a) and ParamReL~(b).}
    \label{fig:param-distribution-structure}
\end{figure}

Following the notations in diffusion models, we use $\mbx_0$ to denote the observations. There are $T$ \emph{reverse} steps in BFNs which gradually {reveals the information of $\mbx_0$} through $\{\mbx_T, \mbx_{T-1}, \ldots, \mbx_1\}$ to the input distribution\footnote{It is noted that the index $t$ is used reversely in \cite{graves2023bayesian}. We make such changes to be consistent with the diffusion model settings~\cite{ho2020denoising,song2020denoising}.}. At the $t$-th step, the reverse step is denoted as a \emph{sender distribution} $p_\mathrm{S}(\mbx_{t-1}\g\mbx_{t};\alpha_t)$, where $\alpha_t$ is a precision parameter. Combined with input distribution $p_\mathrm{I}(\mbx_{t};\mbtheta_{t+1})$, the posterior distribution of $\mbx_{t}$ is obtained as $p(\mbx_{t}; \mbtheta_{t}=h(\mbtheta_{t+1}, \mbx_{t-1}, \alpha_t))\propto p_\mathrm{I}(\mbx_{t};\mbtheta_{t+1})p_\mathrm{S}(\mbx_{t-1}\g\mbx_{t};\alpha_t)$, where $h(\mbtheta_{t+1}, \mbx_{t-1}, \alpha_t)$ is the Bayesian update function. By feeding this posterior parameter $\mbtheta_{t}$ into a neural network $\psi(\cdot)$, $\mbx_t$'s output distribution $p_\mathrm{O}(\cdot)$ is parameterized as $p_\mathrm{O}(\mbx_t;\psi(\mbtheta_{t}))$. Finally, a \emph{receiver distribution} $p_{\mathrm{R}}(\cdot)$ is defined as the expectation of the sender distribution with respect to the output distribution, i.e., $p_{\mathrm{R}}(\mbx_{t-1};\psi(\mbtheta_{t}), \alpha_{t}):=\mathbb{E}_{p_\mathrm{O}(\mbx_t;\psi(\mbtheta_{t}))}[p_\mathrm{S}(\mbx_{t-1}\g\mbx_t;\alpha_t)]$. See Figure~\ref{fig:param-distribution-structure} (a) for a visualization of the relationships between these distributions.

In BFNs, the joint distribution over the observation $\mbx_0$ and the intermediates $\{\mbx_t\}_{t}$ {is defined as} $p(\mbx_0, \{\mbx_t\}_{t}|-):=p_\mathrm{O}(\mbx_0;\psi(\mbtheta_0))\prod_{t=1}^Tp_{\mathrm{R}}(\mbx_{t-1};\psi(\mbtheta_{t}), \alpha_{t})$. This intractable joint distribution can be approximated under the variational inference framework as follows: 
\begin{multline}
\label{eq:bfnLoss}    
    \log p(\mbx_0)
    \ge  \mathbb{E}_{p_\mathrm{F}(\mbtheta_{1:T}|-)p_\mathrm{S}(\{\mbx_t\}_{t}|-)}\left[\log\frac{p_\mathrm{O}(\mbx_0;\psi(\mbtheta_0))\prod_{t=1}^Tp_{\mathrm{R}}(\mbx_{t-1};\psi(\mbtheta_{t}), \alpha_{t})}{\prod_{t=1}^T p_\mathrm{S}(\mbx_{t-1}\g\mbx_t;\alpha_t)}\right] \\
=  -\sum_{t=1}^T \underbrace{\mathbb{E}_{p_F(\mbtheta_{t}|-)}\KL{p_\mathrm{S}\left(\mbx_{t-1} \g{\mbx_0};\alpha_{T:t}\right)}{p_{\mathrm{R}}\left(\mbx_{t-1} ;\psi(\mbtheta_{t}), \alpha_t\right)}}_{\mathcal{L}_t^{\mathrm{R}}(\mbx)} + \underbrace{ \mathbb{E}_{p_F(\mbtheta_{0}|-)}\ln p_\mathrm{O}(\mbx_0;\psi(\mbtheta_0))}_{\mathcal{L}^{\mbD}(\mbx)},
\end{multline}

where $p_\mathrm{F}(\mbtheta_{t}|-)$ represents the distribution of $\mbtheta_t$~(see Appendix~\ref{app:bfn} for a detailed calculation). Maximizing Eq.~\ref{eq:bfnLoss} equals minimizing the discrepancy between the sender and receiver distributions and implicitly penalizing Distortion $\mathcal{L}^{\mbD}(\mbx)$ to maximize the likelihood distribution over data.

By working in the parameter space, BFNs can uniformly model continuous, discretized, and discrete observations. For example, {BFNs can use the mean parameters of Gaussian distributions to model continuous data and the event probability parameters of categorical distributions to study discrete data}~(see detailed settings for distributions in Table~\ref{table:detail-distribution-format}). However, {BFNs cannot produce meaningful latent semantics that discover high-level concepts in observations, such as hair colors in portrait images.}
{\begin{table}
    \centering
    \label{tab:inputDis}
    \caption{Examples of detailed distribution formats in BFNs. }
    \renewcommand{\arraystretch}{1.3}
    \label{table:detail-distribution-format}
    \scalebox{0.9}{
    \begin{tabular}{l||ccc}
    \bottomrule[1pt]
    \textbf{Data type}     & $p_\mathrm{I}(\mbx_t|\mbtheta_{t+1})$ & $p_\mathrm{S}(\mbx_{t-1}|\mbx_t;\alpha_t)$ & $\mbtheta_{t}=h(\mbtheta_{t+1}, \mbx_{t-1}, \alpha_t)$   \\
    \hline
    Continuous data & $\cN(\mbx_t;\mbtheta_{t+1}=\{\mu_{t+1}, \rho_{t+1}^{-1}\})$ & $\cN(\mbx_{t-1};\mbx,\alpha_t^{-1})$ &  $\mu_t = \frac{\alpha_t\mbx_{t-1}+\rho_{t+1}\mu_{t+1}}{\alpha_t+\rho_{t+1}}$   \\
    Discrete data     & $\text{categorical}(\text{x};\frac{\mbone}{K})$ & $\cN(\mbx_{t-1};\alpha_t(K\mbe_{\mbx_{t}}-1), \alpha_t K\mbI)$ & $\mbtheta_t = \frac{e^{\mbx_{t-1}}\mbtheta_{t+1}}{\sum_k e^{\mbx_{t-1,k}}\theta_{t+1,k}}$   \\
    \hline
    \textbf{Data type}      &  $p_\mathrm{O}(\mbx_t|\mbtheta_t)$ & \multicolumn{2}{c}{$p_\mathrm{R}(\mbx_{t-1}|\psi(\mbtheta_{t}), \alpha_t)$}   \\
    \hline
    Continuous data  & $\delta(\mbx_t-\psi(\mbtheta_t))$ &  \multicolumn{2}{c}{$\cN(\mbx_{t-1};\psi(\mbtheta_t),\alpha_t^{-1})$}   \\
     Discrete data      & $\text{categorical}(\text{softmax}(\psi(\mbtheta_t)))$ & 
     \multicolumn{2}{c}{$\sum_kp_O(k;\psi(\mbtheta_t))\cN(\mbx_{t-1};\alpha_t(K\mbe_{k}-1), \alpha_t K\mbI)$}\\
     \bottomrule[1pt]
    \end{tabular}}
\end{table}
}

\section{ParamReL: Parameter Space Representation Learning Framework}
\label{sec:lbfn}
This section introduces the ParamReL framework, which facilitates parameter space representation learning through low-dimensional latent semantics. Instead of approximating the data distribution $p(\mbx_0)$, ParamReL aims to learn the joint distribution over  data $\mbx_0$ and latent semantics $\mbz$, with $|\mbz| \ll |\mbx_0|$. This ensures that low-dimensional latent semantics $\mbz$ are learned from high-dimensional data $\mbx_0$. In other words, ParamReL seeks to reconstruct observations $\mbx_0$ using a high-performing likelihood function while obtaining meaningful low-dimensional latent semantics $\mbz$.

\begin{figure}[!htbp]
    \centering
    \includegraphics[width=1\columnwidth]{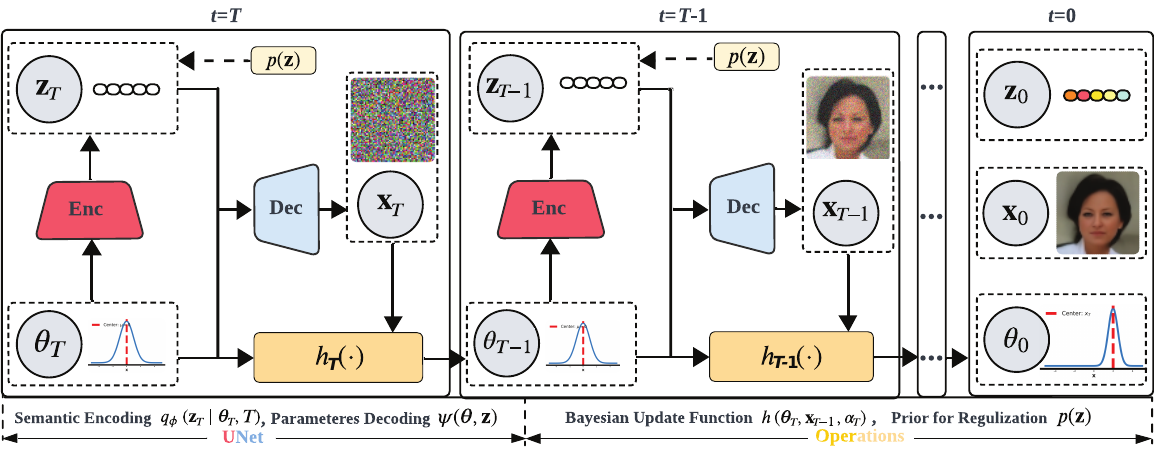}
    \caption{The framework of ParamReL, consisting of a semantic encoder $q_{\phi}$ and a parameter decoder $p_{\Psi}$. During the reverse stage, parameters of data distribution $\theta_{t}$ will encode a time-specific semantic latent $\mbz_{t}$ and then decode the $t-1$-th parameters $\theta_{t-1}$.}
    \label{fig:working-flow-graph}
\end{figure}

Figure~\ref{fig:working-flow-graph} shows the framework of ParamReL, which involves $T$ steps of Bayesian updates, parameter encoding, and latent decoding. At step $t=\{T, T-1, \ldots, 1\}$, the observations first go through a Bayesian update function and a sender distribution to form the posterior 

\subsection{\texorpdfstring{Self-encoder $q_{\mathbf{\phi}}(\mathbf{z}_{t}|\mathbf{\theta}_t, t)$}{Self-encoder q(phi)(z\_t|theta\_t, t)}}

Common autoencoders, such as $q_{\mbphi}(\mbz|\mbx_0)$, aim to encode the observation $\mbx_0$ into a $\emph{static}$ latent semantic $\mbz$. However, this construction does not align well with the ParamReL framework, which employs multiple steps to generate observations. In ParamReL, a single static semantic $\mbz$ may not be sufficient to encode all intermediate variables $\{\mbx_t\}_{t}$. Specifically, $\mbz$ works together with the updated parameter $\mbtheta_t$ in $\psi(\mbtheta_t, \mbz)$. A static semantic $\mbz$ derived from the initial observation $\mbx_0$ may fail to capture the complex dynamics in $\{\mbtheta_t\}_t$. 

In ParamReL, we propose a \emph{self-encoder} $ q_{\mbphi}(\mbz_t|\mbtheta_t, t) $ that differs from the standard encoder $ q_{\mbphi}(\mbz|\mbx_0) $ in two key ways: (1) We adopt a step-wise representation $\mbz_t$ tailored to the variable behaviours at each step $t$; (2) We utilize $\mbtheta_t$, which summarizes information from all previous steps, instead of $\mbx_t$, which only captures the current step $t$, to generate the step-wise $\mbz_t$. Through $ q_{\mbphi}(\mbz_t|\mbtheta_t, t) $, $\mbtheta_t$  effectively encodes \emph{itself} into $\mbz_t$, and they then jointly form $\psi(\mbtheta_t, \mbz_t)$. Therefore, we refer to it as a \emph{self-encoder}. 

The proposed self-encoder works consistently with the decoder network here as both work on $\mbtheta_t$, see Figure~\ref{fig:param-distribution-structure} (b). By obtaining \emph{a series of} latent semantics $\{\mbz_t\}_t$, it is expected these encoded latents would show progressive semantic behaviour (representing, for example, the gradual changes of age \emph{and} smiling \emph{and} skin colour of a person) along with the generation procedures. This is followed by a reconstruction phase, where lower-level generative latent ${\mbx}_{t}$ (such as hair texture) are progressively incorporated. In our experiments, we parameterize $q_{\mbphi}(\mbz_t|\mbtheta_t, t)$ as a UNet encoder (see Appendix \ref{app:enc_fram} for more details).

\textbf{Prior for $\mbz_t$:} Ideally, the latent semantics $\{\mbz_t\}_t$ should provide low-dimensional representations that are different from the intermediates $\{\mbx_{t}\}_t$ of BFNs, yet they should not compromise the data reconstruction process. To generate high-quality semantic representations, a learnable and smooth latent space is essential~\cite{bengio2013representation}. This necessitates the integration of the prior distribution $p(\mbz_t)$ into a robust probabilistic framework, thereby enabling the unconditional sampling of ${\mbx}_0$. For simplicity and sampling efficiency, $p(\mbz_t)$ could be assumed to follow a straightforward distribution, such as a Gaussian.

\subsection{Training and Testing with ParamReL}
\label{sec:rlbfn}
By augmenting the latents $\{\mbz_t\}_t$, the joint distribution of $\mbx_{0}, \{{\mbx}_t\}_{t}, \{\mbz_t\}_t$ can be written as:
\begin{align}\label{eq:joint_ParamReL}
        p\left({\mbx}_{0}, \{{\mbx}_t\}_{t}, \{\mbz_t\}_t\g\mbtheta_0, \alpha\right)=  p_\mathrm{O}(\mbx_0;\psi(\mbtheta_0, \mbz_0))\prod_{t=1}^Tp(\mbz_t)\mathbb{E}_{p_\mathrm{O}(\mbx_t;\psi(\mbtheta_t, \mbz_t))}[p_\mathrm{S}(\mbx_{t-1}\g\mbx_t;\alpha_t)],
\end{align}
where the model implements an output distribution $p_\mathrm{O}(\mbx_t;\psi(\mbtheta_t, \mbz_t))$ conditioned on auxiliary latent ${\mbz}_t$, which is distributed according to prior $p({\mbz}_t)$. The latent variable ${\mbz}_t$ of prior $p({\mbz}_t)$ is independent of the BFNs process as a latent representation of input. {Through a neural network architecture $\psi(\cdot, \cdot)$, $\mbz_t$ works together with the posterior parameter $\mbtheta_t$ to form the parameters of the output distribution as $\psi(\mbtheta_t, \mbz_t)$.} We use $p_\mathrm{O}(\mbx_0;\psi(\mbtheta_0, \mbz))\prod_{t=1}^Tp_{\mathrm{R}}(\mbx_{t-1};\psi(\mbtheta_t, \mbz_t), \alpha_t)$ to decode {both the auxiliary latent representation and the generative latent}.

\textbf{Training with ParamRL:} Variational inference methods approximate the intractable joint distribution. With $q_{\mbphi}(\mbz_t\g\mbtheta_t, t)$ defined as the encoder for $\mbz_t$ and $p_\mathrm{S}(\{\mbx_t\}_{t}|-)$ defined as the variational distribution for $\mbx_{T:1}$, the evidence lower bound on the marginal log-likelihood of data $\mbx_0$ is (see the full derivation in Appendix \ref{app:der_ParamReL}): 
{\begin{multline}    
\label{eq:ParamReL-elbo}
\log p(\mbx_{0}) 
 \geq -\sum_{t=1}^T \mathbb{E}_{p_\mathrm{F}(\mbtheta_{t}|-)}\mathbb{E}_{q_{\mbphi}(\mbz_t)}\left\{{\KL{p_\mathrm{S}\left(\mbx_{t-1} \g{\mbx_0};\alpha_{T:t}\right)}{p_{\mathrm{R}}\left(\mbx_{t-1} ;\psi(\mbtheta_t, \mbz_t), \alpha_t\right)}}\right.  \\
 \left.- \KL{q_{\mbphi}\left({\mbz}_t \g {\mbtheta_t}, t\right)}{p({\mbz}_t)}\right\}+ \mathbb{E}_{q_{\mbphi}(\mbz_0,\mbtheta_0)}\left[{\ln p_\mathrm{O}(\mbx_0;\psi(\mbtheta_0, \mbz_0))}\right]:=\mathcal{L}_{\texttt{ParamReL}}.
\end{multline}}

Maximizing the objective function $\cL_{\text{ParamReL}}$ is equivalent to performing amortized inference ~\cite{kingma2013auto} through encoders $q_{\mbphi}(\mbz_t|\mbtheta_t, t)$ and learning likelihood function through decoders~\cite{zhao2017infovae}. The encodable posterior $q_{\mbphi}\left(\mbz_t|\mbtheta_t,t\right)$ is used to infer high-level semantics $\{{\mbx}_t\}_{t}$. During the \emph{decoding} stage, {${\mbx}_0$ is used into generative latent variables $\mbx_{1:T}$}, which contains low-level semantic features in generating the observations. {In ParamReL, the parameters of the output distribution are learned through iteratively proceeding the Bayesian updating process and a learned noise model ${\psi}({\mbtheta}, \mbz)$ parameterized by neural networks $\psi$.}

\textbf{Sampling and Reverse-sampling with ParamReL in the Test Phase} Once ParamReL is trained, we can use the generative procedure to do pseudo-data sampling, i.e., use $\mbx_T$ to sequentially generate $\mbx_{T-1}, \dots, \mbx_0$ from $\mbx_T$. In particular, at the $t$-th step, we have
\begin{equation}
\begin{aligned}
\widehat{\mbx}_{t-1}\sim  p_{\mathrm{S}}\left(\widehat{\mbx}_{t-1} |\mbx_t, \alpha_t\right), \ 
\mbtheta_t = h(\mbtheta_{t+1},\widehat{\mbx}_{t-1},\alpha_t), \ 
\mbz_t\sim p(\mbz_t), \  
\widehat{\mbx}_{t}\sim p_O(\widehat{\mbx}_{t};\psi(\mbtheta_t, \mbz_t)), \ 
\mbx_{t-1}=\widehat{\mbx}_t
\end{aligned}
\end{equation}

where $\widehat{\mbx}_{t-1}$ is a pseudo value for the intermediate $\mbx_{t-1}$. Since the self-encoder is already trained, it may be possible to use $q_{\mbphi}(\mbz_t|\mbtheta_t, t)$ to replace $\mbz_t$'s prior $p(\mbz_t)$ to improve the sampling quality. 

However, the reverse-sampling procedure, which transits observation $\mbx_0$ through $\mbx_1, \mbx_2, \ldots$ until its noisy state $\mbx_T$, is not as straightforward as the sampling procedure. In fact, we notice the output distribution $p_O(\mbx_t;\psi(\mbtheta_t, \mbz_t))$ may help us to derive the reverse-sampling procedure. {At the $t$-th step, the reverse-sampling procedure may be implemented as
\begin{align}
    \mbtheta_{t+1} = h^{-1}(\mbtheta_{t}, \mbx_{t-1}, \alpha_t), \quad \mbz_{t+1} = q_{\mbphi}(\mbtheta_{t+1}, t+1), \quad \mbx_{t+1}\sim p_\mathrm{O}(\mbx_{t+1};\psi(\mbtheta_{t+1}, \mbz_{t+1})).
\end{align}}
where $h^{-1}(\cdot)$ refers to the inverse Bayesian update function. Given the sampling and reverse-sampling procedure, we can do various tasks such as image reconstruction, interpolation, etc. The details of such results are shown in the experiments.

\subsection{Regularizing Semantics By Maximizing Mutual Information}
Ideally, during the training phase, {we want to acquire the semantic representation $\mbz_{t}$ by the self-encoder $q_{\mbphi}(\mbz_t|\mbtheta_t, t)$ and achieve high-quality reconstruction $\widehat{\mbx_{0}}$ by the decoder}~(i.e., the output distribution $p_O(\mbx_0;\psi(\mbtheta_0, \mbz_0))$). However, there exists a trade-off between inference and learning~\cite{shao2020controlvae, wu2023evae} coherent in optimizing the objective function $\mathcal{L}_{\text{ParamReL}}$ in Eq.~(\ref{eq:ParamReL-elbo}). In most cases, ParamReL favours fitting likelihood rather than inference~\cite{zhao2019infovae}. {Based on the rate-distortion theory~\cite{alemi2018fixing,bae2022multi}, the rate (KL divergence term) constrained by the encoders compresses enough information for the distortion (reconstruction error) but less informative to induce a smooth latent space.} 


To remedy the insufficient representation learning during the inference stage, we want to increase the dependence between input parameters $\mbtheta_t$ and encoded latent $\mbz_t$ by maximizing the mutual information $I(\mbtheta_t, \mbz_t)$. 

We can rewrite the tractable learning object in ParamReL by adding the mutual information maximization term as: $\mathcal{L}_{\text{ParamReL+}}=\mathcal{L}_{\text{ParamReL}}+\frac{\gamma}{T} \sum_tI_q(\mbtheta_t; \mbz_t)$, where $I_q(\mbtheta_t; \mbz_t)$ is the mutual information between $\mbtheta_t,\mbz_t$ under the distribution $q_{\phi}(\mbz_t|\mbtheta_t)$. Considering that we cannot optimize this object directly, we can rewrite this object by factorizing the rate term into MI and Total Correlation (TC): 

\begin{multline}    
\mathcal{L}_{\texttt{ParamReL+}}= -\sum_{t=1}^T \mathbb{E}_{p_\mathrm{F}(\mbtheta_{t}|-)}\mathbb{E}_{q_{\mbphi}(\mbz_t)}\left\{{\KL{p_\mathrm{S}\left(\mbx_{t-1} \g{\mbx_0};\alpha_{T:t}\right)}{p_{\mathrm{R}}\left(\mbx_{t-1} ;\psi(\mbtheta_t, \mbz_t), \alpha_t\right)}}\right.  \\
 \left.- \frac{1-\gamma}{T}\KL{q_{\mbphi}\left({\mbz}_t \g {\mbtheta_t}\right)}{p({\mbz})}- \frac{\gamma+\lambda-1}{T}\KL{q_{\mbphi}\left({\mbz}_t\right)}{p({\mbz})}\right\}+ \mathbb{E}_{q_{\mbphi}(\mbz_0, \mbtheta_0)}\left[{\ln p_\mathrm{O}(\mbx_0;\psi(\mbtheta_0, \mbz_0))}\right].
\label{eq:ParamReL-elbo_mi}
\end{multline}
\textbf{Mutual Information Learning}
Unlike the rest of the terms that can be optimized directly using reparameterization tricks, the TC term cannot be directly optimized due to intractable marginal distribution $q_{\mbphi}({\mbz_t})$. Here, we follow the guidance in ~\cite{zhao2019infovae} to replace the TC term with any strict divergence $D$, where $D\left(q_{\mbphi}(\mbz) \| p(\mbz)\right)=0$ iff $q_{\mbphi}(\mbz)=p(\mbz)$. We implement the Maximum-Mean Discrepancy (MMD)~\cite{zhao2019infovae} from the divergence family. MMD is a statistical measure that quantifies the difference between two probability distributions by comparing their mean embeddings in a high-dimensional feature space. By defining the kernel function $\kappa(\cdot, \cdot)$, $D_{\mathrm{MMD}}$ is denoted as:
\begin{equation}
D_{\mathrm{MMD}}\left(q(\cdot) \| p(\cdot)\right)  =\mathbb{E}_{p(\mbz), p\left(\mbz^{\prime}\right)}\left[\kappa\left(\mbz, \mbz^{\prime}\right)\right]-2 \mathbb{E}_{q(\mbz), p\left(\mbz^{\prime}\right)}\left[\kappa\left(\mbz, \mbz^{\prime}\right)\right] +\mathbb{E}_{q(\mbz), q\left(\mbz^{\prime}\right)}\left[\kappa\left(\mbz, \mbz^{\prime}\right)\right].
\end{equation}

\section{Related Work}

{Recent advances have demonstrated that diffusion-based models \cite{ho2020denoising, song2020denoising} are capable of generating high-quality data. Nonetheless, compared to the autoencoder framework, the intermediate outputs in diffusion stages are high-dimensional and lack smoothness, making them unsuitable for representation learning. Contemporary research focuses on encoding a conditional latent space to acquire semantic low-dimensional representations. However, sample-based models \cite{preechakul2022diffusion, wang2023infodiffusion}, such as VAEs and Diffusions, exhibit limitations when applied to discrete data.

 Deep hierarchical VAEs have seen progress in capturing latent dependence structures for encoding an expressive posterior, statistically or semantically. VQVAE-based 
~\cite{van2017neural,razavi2019generating} models have local-to-global feature-based explanatory hierarchies at the image level, forming a codebook-based discrete posterior. \cite{sonderby2016ladder,tomczak2018vae} build recursive latent structures in multi-layer networks to form an aggregated posterior. NVAE~\cite{vahdat2020nvae} demonstrates that depth-wise hierarchies encoded by residual networks can approximate the posterior precisely despite using shallow networks. Unlike the sample-based encoder, where the information flow between input and latent is maximized in encoding-decoding pipelines on the sample space, we develop progressive encoders in the parameter space to capture the dynamic semantics.}

\section{Experiments}

We present two ParamReL variants in different parameter spaces: for the continuous input distribution, denoted as ParamReLc, and for the discrete input distribution, denoted as ParamReLd; refer to Table ~\ref{tab:inputDis} for the distribution definitions. We evaluate ParamReLs on discrete and continuous benchmarks, including three discrete datasets binarized by grey-style data: binarized MNIST (bMNIST) \cite{deng2012mnist}, binarized FashionMNIST (bFashionMNIST) ~\cite{xiao2017fashion}; as well as three continuous datasets: CelebA ~\cite{liu2015deep}, CIFAR10 ~\cite{krizhevsky2009learning}, and Shapes3d ~\cite{3dshapes18} \footnote{For the discrete version, continuous data ($k$-bit images) can be discretized into $2^k$ bins. This process involves dividing the data range $[-1, 1]$ into $k$ intervals, each with a length of $2/k$.}.The hyperparameter choices and experiment configurations corresponding to each dataset for training can be found in Appendix~\ref{app:hyer}. 

We evaluate the representation learning ability in three reconstruction-based tasks, i.e., decoding the samples from the \textit{encoder and the decoder}, including latent interpolation, disentanglement, and time-varying conditional reconstruction; and one generation-based task, i.e., decoding the samples \textit{only from the decoder} with a given prior, including conditional generation.

We conduct a comprehensive two-fold comparison. Firstly, we compare our parameter-based variants against sample-based representation learning baselines, including AE and VAE-based models: $\beta$-VAE \cite{higgins2017beta}, infoVAE \cite{zhao2017infovae}, and diffusion-based models: DiffusionAE \cite{preechakul2022diffusion}, infoDiffusion \cite{wang2023infodiffusion}. These models represent significant advancements in representation learning, with $\beta$-VAE being the pioneer in VAE-based disentanglement, infoVAE introducing MMD to strike a balance between generation and representation, DiffusionAE integrating AEs into Diffusion models for encodable latent learning, and infoDiffusion incorporating VAEs into diffusion models for disentanglement learning. Secondly, we compare the ParamReLd and ParamReLc variants across various input distributions for continuous and discrete data representation learning. This comparison enables a detailed investigation of parameter space assumptions for discrete and continuous representation learning.

\label{sec:exp}

\subsection{Representation Learning on Discrete Data by ParamReLd}

\begin{figure}[!htbp]
    \centering
    \includegraphics[width=1\columnwidth]{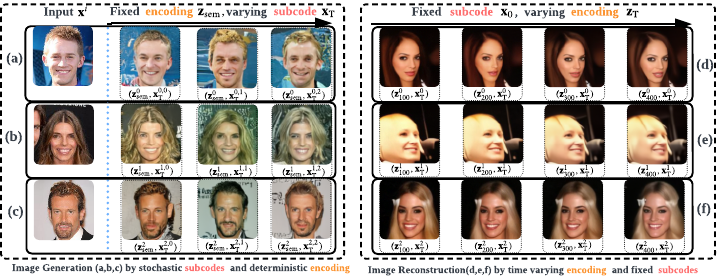}
    \caption{The left panel~(a-b) shows high-level semantic latent captured by $\mbz_{\mathrm{sem}}$ from ParamReL's encoders. By fixing $\mbz_{\mathrm{sem}}$, the global characters of the images are invariant. By varying the stochastic subcodes $\mbx_{0}$, the local attributes in the corresponding generated images may vary, such as the \texttt{Narrow\_Eyes} attribute in (a), the \texttt{Blond\_Hair} attribute in (b), and the \texttt{Mouth\_Slightly\_Open} attribute in (c). The right panel~(d-f) illustrates the time-varying changes that ParamReL's progressive encodes interfaced. By varying time encodes at 100,200,300 time steps, more attributes will be influenced in the reconstruction stage: the \texttt{Big\_Lips, Pointy\_Nose} attributes in (d), the \texttt{Blond\_Hair, Bald} attributes in (e) and the \texttt{Wavy\_Hair, High\_Cheekbones} attributes in (f). }
    \label{fig:lowlevel-z}
\end{figure}

\subsubsection{Semantic Latent for Downstream Tasks}
The representation $\mbz_{\mathrm{sem}}$ should be \textit{general} and \textit{transferable} ~\cite{franceschi2019unsupervised} for downstream tasks. We measure the quality of learned representations in classification tasks. To ensure their universality, we assess the quality of the learned representations on various datasets by different classifiers. Specifically, Following~\cite{xiao2023trading}, we train a classifier on labelled test sets for each ParamReL model. We allocate $80\%$ of the dataset for training a classifier and reserve the remaining $20\%$ for testing purposes. The performance on the test set is evaluated based on AUROC. This process is conducted in a 5-fold cross-validation manner, with the results reported as mean metrics ± one standard deviation. The results are shown in Table \ref{tab:zclass}. Higher AUROC would suggest that the learned features $\mbz_{\mathrm{sem}}$ contain more information about data $\mbx$. In addition to assessing the representation quality, we also compare the image reconstruction ability against baselines. From Table \ref{tab:zclass}, we can conclude that VAE-based models still produce blurry reconstructions, while diffusion-based and parameter-based models can build near-exact reconstructions. Refer to Figure \ref{fig:recon-bfn-mnist} in Appendix \ref{app:recon} for reconstructed binary images.

\subsection{Representation Learning on Continuous Data by ParamReLs}

\subsubsection{High-level Representation Learning for Conditional Generation}
In Figure \ref{fig:lowlevel-z}, it is demonstrated that high-level semantic information is captured by the auxiliary variables $\mbz_{\mathrm{sem}}$ in image generation tasks. This is illustrated by a set of latent variables $<\mathbf{z}_{\mathrm{sem}}^{i}, \mathbf{x}_{\mathrm{0}}^{i,j}>$, where the auxiliary variables are fixed and encoded from the $i$-th input image by trained the ParamReL's encoders. Additionally, stochastic subcodes $\mbx_0$ are sampled $j$ times from $\mathcal{N}(0,1)$ corresponding to the $i$-th input image. Concurrently, the low-level semantic features, such as local attributes in images (e.g., \texttt{Wearing\_Hat}, \texttt{Big\_Nose}), are captured by stochastic subcodes in the BFN decoders of ParamReL.

\subsubsection{Time-varying Representation Learning for Conditional Reconstruction}
The effectiveness of progressive latent learned by the \emph{self-encoder} will be validated on a new time-varying reconstruction task. Given input $\mbx^{i}$, the time encoding latent set will be acquired by encoding the $t$-th time and parameters, i.e., $\{\mbz_{0},...,\mbz_{1000}\} \sim q_{\mbphi}(\mbz_t|\mbtheta_t, 
 t)$. We will reconstruct samples with time-specific semantic encoding and sample-specific subcodes, along with the fixed subcode reversed from the same input $\mbx^{i}$. In that case, the attributes will vary due to the semantics evolution encoded by time-specific latent. Refer to Figure \ref{fig:lowlevel-z} for more explanation.
\begin{table}[ht]
\centering
\caption{Comparison of representation learning algorithms on discrete data by classification accuracy and generation performance. ($\uparrow$ higher is better; $\mathit{g}$ for the Gaussian distribution, $\mathit{c}$ for the categorical distribution; [\textcolor{myblue}{\textbf{Top-1}},\textcolor{myorange}{\textbf{Top-2}}], \textbf{S} for sample-based models, \textbf{P} for parameter-based models).}
\newcommand{\greencheck}{{\bf \color{OliveGreen}\checkmark}}
    \newcommand*\colourcheck[1]{%
      \expandafter\newcommand\csname #1check\endcsname{\textcolor{#1}{\ding{52}}}%
    }
    \definecolor{bloodred}{HTML}{B00000}
    \definecolor{cautionyellow}{HTML}{EED202}
    \newcommand*\colourxmark[1]{%
      \expandafter\newcommand\csname #1xmark\endcsname{\textcolor{#1}{\ding{54}}}%
    }
    \newcommand*\colourcheckodd[1]{%
      \expandafter\newcommand\csname #1checkodd\endcsname{\textcolor{#1}{\ding{51}}}%
    }
    \colourcheckodd{cautionyellow}
    \colourcheck{cautionyellow}
    \colourcheck{OliveGreen}
    \colourxmark{bloodred}
\newcommand{\ourxmark}{\bloodredxmark}%
\newcommand{\ourcheckmark}{\OliveGreencheck}
\label{tab:zclass}
\scriptsize
\begin{tabularx}{\textwidth}{>{\hsize=0.5\hsize}X>{\hsize=0.5\hsize}X>{\hsize=2\hsize}X*{6}{>{\centering\arraybackslash\hsize=1\hsize}X}}
\toprule[1pt]
 &   && \multicolumn{2}{c}{\textbf{bMNIST}} & \multicolumn{2}{c}{\textbf{bFASHIONMNIST}}\\
\cmidrule(lr){4-5} \cmidrule(lr){6-7} \cmidrule(lr){8-9}
 & \textbf{Prior}&  & \textbf{AUROC}$\uparrow$ & \textbf{NotBlur?} & \textbf{AUROC}$\uparrow$ & \textbf{NotBlur?} \\
\hline
\multirow{6}{*}{\textbf{S}} 
&-&\textbf{AE} & - & \ourxmark & 0.819\tiny±0.003 &  \ourxmark \\
&$\mathit{g}$&\textbf{VAE} &- &\ourxmark  & 0.796\tiny±0.002 & \ourxmark \\
&$\mathit{g}$& $\beta$-\textbf{VAE}  &0.842\tiny±0.459  &\ourxmark & 0.779\tiny±0.004 & \ourxmark\\
&$\mathit{g}$&\textbf{infoVAE} &0.847\tiny±0.386 & \ourxmark & 0.807\tiny±0.003& \ourxmark\\
&$\mathit{g}$&\textbf{DiffAE}  & - & \ourcheckmark &0.835\tiny±0.002 & \ourcheckmark\\
&$\mathit{g}$&\textbf{infoDiffusion}  &0.898\tiny±0.430 & \ourcheckmark &0.839\tiny±0.003 & \ourcheckmark\\
\hline
\multirow{2}{*}{\textbf{P}} 
&$\mathit{c}$&\textbf{ParamReLd}\tiny($\lambda=0.1$)  &\textcolor{myorange}{\textbf{0.927\tiny±0.086}} &\ourcheckmark &\textcolor{myorange}{\textbf{0.857\tiny±0.666}} & \ourcheckmark \\
&$\mathit{c}$&\textbf{ParamReLd} \tiny(w ProgEnc)  &\textcolor{myblue}{\textbf{0.946\tiny±0.334}} & \ourcheckmark&\textcolor{myblue}{\textbf{0.892\tiny±0.787}} & \ourcheckmark\\
\bottomrule[1pt]
\end{tabularx}
\end{table}

\subsubsection{Smooth Representation Learning for Latent Interpolation}

Latent space interpolation ~\cite{goodfellow2014generative,higgins2017beta} can validate the smoothness, continuity, and semantic coherence in the learned representation learning space of generative models. Generally, we can embed two samples into the corresponding latent space, and interpolating between latent variables sampled from the corresponding space yields interpolated representations. Their reconstruction learned by generative models reveals the semantic richness of the learned space. The specific latent interpolation process is detailed in Appendix~\ref{app:inter}.

In Figure \ref{fig:interpolation},  in contrast to VAE variants (a) vanilla VAE, and (b) $\beta$-VAE, ParamReL allows near-exact reconstruction. Compared with diffusion models (c) DiffusionAE and (d) infoDiffusion, ParamReL has a smoother and more consistent latent space with high-quality samples.
\begin{figure}[!htbp]
    \centering
    \includegraphics[width=1\columnwidth]{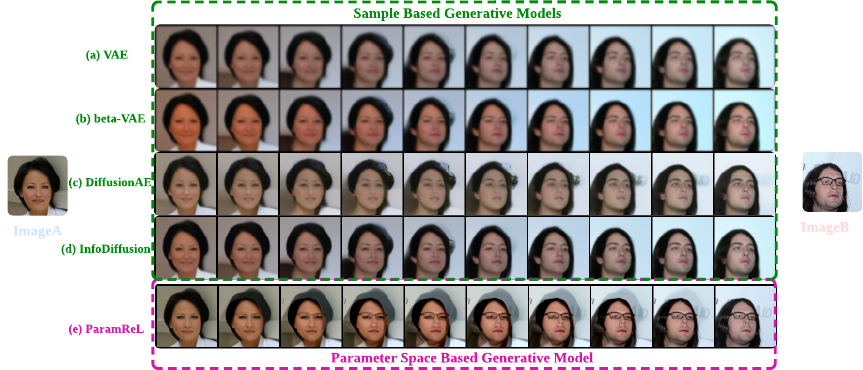}
    \caption{Comparisons of  latent space interpolation among \textcolor{mygreen}{sample-based models} and \textcolor{mypurple}{parameter-based models} on dataset CelebA. Only our ParamRel model (e) can learn a continuous, smooth latent space while ensuring near-exact image reconstruction. Specifically, while sample-based generative models can learn a continuous but unsmooth latent space, this leads to incomplete reconstructions. For example, in (a-d), the attribute of eyeglasses is frequently omitted. Moreover, VAEs (a,b) tend to produce blurry images. Additionally, it is observable that sample-based models often compromise reconstruction in favour of representation learning, as evidenced by the failure of diffusion model variants (c-d) to accurately reconstruct background characters in \textcolor{mypink}{imageB}. }
    \label{fig:interpolation}
\end{figure}

\subsubsection{Disentanglement}
\begin{figure}[!htbp]
    \centering
    \includegraphics[width=1\columnwidth]{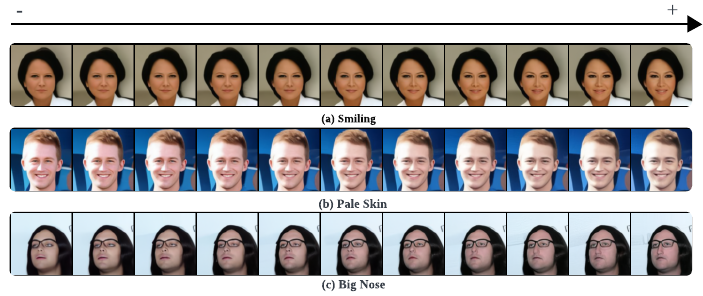}
    \caption{Traversals of latent by ParamReL on CelebA. The interpretable traversal directions are displayed by traversing the encodings ranging from $[-3,3]$.}
    \label{fig:disentanglement}
\end{figure}

We conduct latent traversals on the CeleA dataset to assess the disentanglement properties of our trained ParamReL, as depicted in Figure \ref{fig:disentanglement}. Specifically, we modify one dimension at a time, i.e., $d (0 \le d \le L-1)$, within the semantic latent vector $\mbz_{\mathrm{sem}} \in \mathbb{R}^{L}$, and then varying it $M$ times in a standardized range (e.g., -3 to +3) to acquire traversed latent $\mbz_{\mathrm{traversal}}\in \mathbb{R}^{M \times L}$, while keeping other dimensions constant. Upon decoding these adjusted traversed latent variables, the generated samples are evaluated for changes in specific attributes. Successful disentanglement is demonstrated when the manipulation of one latent dimension changes a single distinguishable attribute in the dataset, such as age, with all other attributes remaining unchanged. This method provides a straightforward validation of the model’s disentanglement capabilities.

Figure \ref{fig:disentanglement} reveals that ParamReL can isolate and control data attributes in an unsupervised manner. This independence in attribute variation underscores the effectiveness of a model in achieving disentanglement. Specifically, the reconstructed sample changes independently in some semantic attribute, e.g., \texttt{Smiling} in Figure~\ref{fig:disentanglement} (a), \texttt{Pale Skin} in Figure~\ref{fig:disentanglement} (b), \texttt{Big Nose} in Figure~\ref{fig:disentanglement} (c) without affecting others. 

To facilitate a thorough and unbiased quantitative assessment of disentanglement, we employ two metrics: a prediction-based indicator: Disentanglement, Completeness, and Informativeness (DCI)~\cite{eastwood2018framework}, and an intervention-based criterion: Total AUROC Difference (TAD)~\cite{yeats2022nashae}. Meanwhile, we report the reconstruction quality in Appendix~\ref{app:recon} and conclude that ParamReL can achieve near-exact reconstruction on Celeba (Figure~\ref{fig:recon-bfn-celeba}), 3DShapes (Figure~\ref{fig:recon-bfn-3dshapes}) and CIFAR10 (Figure~\ref{fig:recon-bfn-cifar10}).

\begin{table}[ht]
\centering
\caption{Comparison of representation learning algorithms on continuous data by disentanglement performance (mean ± std) and classification. The quantitative results for each algorithm averaged over five trials. ($\mathit{g}$ for the Gaussian distribution, $\mathit{c}$ for the categorical distribution, and $\mathit{d}$ for the delta distribution; $\uparrow$ higher is better; [\textcolor{myblue}{\textbf{Top-1}},\textcolor{myorange}{\textbf{Top-2}}]).}
\newcommand{\greencheck}{{\bf \color{OliveGreen}\checkmark}}
    \newcommand*\colourcheck[1]{%
      \expandafter\newcommand\csname #1check\endcsname{\textcolor{#1}{\ding{52}}}%
    }
    \definecolor{bloodred}{HTML}{B00000}
    \definecolor{cautionyellow}{HTML}{EED202}
    \newcommand*\colourxmark[1]{%
      \expandafter\newcommand\csname #1xmark\endcsname{\textcolor{#1}{\ding{54}}}%
    }
    \newcommand*\colourcheckodd[1]{%
      \expandafter\newcommand\csname #1checkodd\endcsname{\textcolor{#1}{\ding{51}}}%
    }
    \colourcheckodd{cautionyellow}
    \colourcheck{cautionyellow}
    \colourcheck{OliveGreen}
    \colourxmark{bloodred}
\newcommand{\ourxmark}{\bloodredxmark}%
\newcommand{\ourcheckmark}{\OliveGreencheck}
\label{tab:disentanglement}
\scriptsize
\resizebox{\textwidth}{!}{
\begin{tabularx}{\textwidth}{>{\bfseries}cc>{\hsize=2.5\hsize}X>{\hsize=1.5\hsize}X*{2}{>{\centering\arraybackslash\hsize=0.4\hsize}X} @{\hspace{10pt}} *{5}{>{\centering\arraybackslash}X}}
\toprule[1pt]
\textbf{} & \textbf{} & \textbf{}& \multicolumn{4}{c}{\textbf{CelebA}} & \multicolumn{2}{c}{\textbf{3DShapes}} \\
\cmidrule(lr){4-7} \cmidrule(lr){8-9}
\textbf{} & \textbf{Prior} & \textbf{} & \textbf{TAD}$\uparrow$ & \textbf{ATTRS}$\uparrow$ & \textbf{NotBlur?} & \textbf{AUROC}$\uparrow$ & \textbf{DCI}$\uparrow$ & \textbf{NotBlur?}  \\
\hline
\multirow{6}{*}{\textbf{S}} 
&- & \textbf{AE} & 0.042\tiny±0.004 & 1.0\tiny±0.0 & \ourxmark &  0.759\tiny±0.003 &  0.219\tiny±0.001 & \ourxmark  \\
&$\mathit{g}$  & \textbf{VAE} & 0.000\tiny±0.000 & 0.0\tiny±0.0 &  \ourxmark & 0.770\tiny±0.002 &0.276\tiny±0.001  & \ourxmark \\

&$\mathit{g}$  & \textbf{$\beta$-VAE} &  0.088\tiny±0.051& 1.6\tiny±0.8 & \ourxmark & 0.699\tiny±0.001 &0.281\tiny±0.001  & \ourxmark  \\
&$\mathit{g}$  & \textbf{infoVAE} & 0.000\tiny±0.000 & 0.0\tiny±0.0 & \ourxmark  & 0.757\tiny±0.003 &0.134\tiny±0.001 & \ourxmark  \\
&$\mathit{g}$ & \textbf{DiffAE} &  0.155\tiny±0.010 & 2.0\tiny±0.0 &    \ourcheckmark &0.799\tiny±0.002  &0.196\tiny±0.001 & \ourcheckmark   \\

&$\mathit{g}$ &\textbf{infoDiffusion} & 0.299\tiny±0.006 & 3.0\tiny±0.0  & \ourcheckmark  & 0.848\tiny±0.001 &0.342\tiny±0.002 & \ourcheckmark   \\
\hline
\multirow{1}{*}{\textbf{P}} 
&$\mathit{c}$& \textbf{ParamReL}\tiny($\lambda =0.1$)  & \textbf{0.261\tiny±0.01}& \textcolor{myblue}{\textbf{5.0\tiny±0.0}} & \ourcheckmark & 0.846\tiny±0.349 & 0.477\tiny±0.052 & \ourcheckmark  \\
\hline
\multirow{2}{*}{\textbf{P}}
&$\mathit{d}$& \textbf{ParamReL}\tiny($\lambda =0.01$)& \textcolor{myorange}{\textbf{0.302\tiny±0.005}}& \textcolor{myorange}{\textbf{4.0\tiny±0.0}} & \ourcheckmark & \textcolor{myorange}{\textbf{0.850\tiny±0.116}} & \textcolor{myblue}{\textbf{0.567\tiny±0.005}}  & \ourcheckmark  \\
&$\mathit{d}$& \textbf{ParamReL}\tiny($\lambda =0.1$)  &  \textcolor{myblue}{\textbf{0.368\tiny±0.005}}& 3.0\tiny±0.0  & \ourcheckmark & \textcolor{myblue}{\textbf{0.865\tiny±0.064}} &\textcolor{myorange}{\textbf{0.485\tiny±0.041}} &  \ourcheckmark \\
\bottomrule[1pt]
\end{tabularx}
}
\end{table}

The qualitative latent traversal and quantitative disentanglement metrics show that learning with ParamReL leads to a visual traversal, which intuitively matches the attribute on which the latent representation is an effective detector.

\section{Conclusion and Limitations}
{A novel parameter space representation learning framework ParamReL is introduced in this work, which provides a uniform strategy for dealing with continuous, discretized, and discrete data. Unlike previous encoder methods that encode observations into static latent semantics, a self-encoder is capable of obtaining a series of progressively structured latent semantics from the step-wise parameters. Experimental results on tasks of latent interpolation, disentanglement, time-varying conditional reconstruction, and conditional generation validate the merits of ParamReL, demonstrating its superior performance in gaining unified representations and a clearer visual understanding.}

ParamReL discloses interesting follow-ups, including: (1) The precision variables $\alpha_t$ could potentially be optimized to accelerate the sampling procedure, which will be further explored. 
(2) Utilizing a pre-trained model for the U-Net architecture might enhance {the performance of the proposed ParamReL}, which will be investigated in future work.

\bibliographystyle{plain}

\newpage 
\appendix
\section{Preliminaries}
\subsection{Bayesian Flow Distribution}
\label{app:bfn}
Bayesian flow distribution $p_\mathrm{F}(\cdot \mid \mbx; t)$ is the marginal distribution over input parameters at time $t$, given prior distribution, accuracy schedule $\alpha$ and Bayesian update distribution $p_U(\cdot \mid \mbtheta, \mbx ; \alpha)$, as follows:

\begin{equation}
p_\mathrm{F}(\mbtheta \mid \mbx ; t)=p_U\left(\mbtheta \mid \mbtheta_0, \mbx ; \beta(t)\right) .
\end{equation}

\label{sec:pre}
\subsection{Generative Latent Variable Models for Representation Learning}
Latent Variable Models~(LVMs)~\cite{everett2013introduction} which aim at learning the joint distribution $p(\mbx,{\mbz})$ over data $\mbx$ and latent variables ${\mbz}$ present efficient ways for uncovering hidden semantics. In LVMs, the joint distribution $p(\mbx, \mbz)$ is usually decomposed as
$p(\mbx,\mbz)= p(\mbx \mid {\mbz}) p({\mbz})$, where $p({\mbz})$ represents prior knowledge for inference~\cite{tschannen2018recent}, thus facilitating learning the conditional distribution $p(\mbx \mid {\mbz})$.  
Among LVMs, Variational AutoEncoders~(VAEs)~\cite{kingma2013auto} and diffusion models~\cite{ho2020denoising,song2020denoising} are two representative approaches~\cite{kwok2012priors}. 

In VAEs, latent variables $\mbz$ is obtained through an \emph{encoder network} $q_{\mbphi}(\mbz\g\mbx)$, whereas observations are reconstructed through a \emph{decoder network} $p_{\mbtheta}(\mbx\g\mbz)$, with $\mbphi$ and $\mbtheta$ being the encoder and decoder parameters. 
The dimensions of $\mbz$ are usually much smaller than those of $\mbx$, denoted as $|\mbz|\ll|\mbx|$, such that redundant information is effectively removed and the most semantically meaningful factors are abstracted~\cite{louizos2015variational}. VAEs are popular for downstream tasks like disentanglement ~\cite{higgins2017beta,yang2021towards,hwang2023maganet,esmaeili2024topological}, classification ~\cite{takahashi2022learning,tonekaboni2022decoupling}, and clustering~\cite{jiang2016variational,xu2021multi}.

On the other hand, diffusion models~\cite{ho2020denoising,song2020denoising} first use $T$ diffusion steps to transform observation $\mbx$ into a white noise $\mbx_T$ and then use $T$ denoising steps to reconstruct the observation. Diffusion models have obtained impressive performance in the fidelity and diversity of generation tasks. However, they might be unable to obtain meaningful latent semantics since the dimensions of $\mbx$ and $\mbx_T$ are the same as $|\mbx|=|\mbx_T|$.  
~\cite{preechakul2022diffusion,wang2023infodiffusion} have attempted to integrate a decodable auxiliary variable $\mbz$ to enable diffusion models to obtain low-dimensional latent semantics. However, they have not overcome issues like the slow training speed inherent to the diffusion and reverse processes.

\subsection{Illustration of Parameter Space Optimization}
\begin{figure}[!htbp]
    \centering
    \includegraphics[width=1\columnwidth]{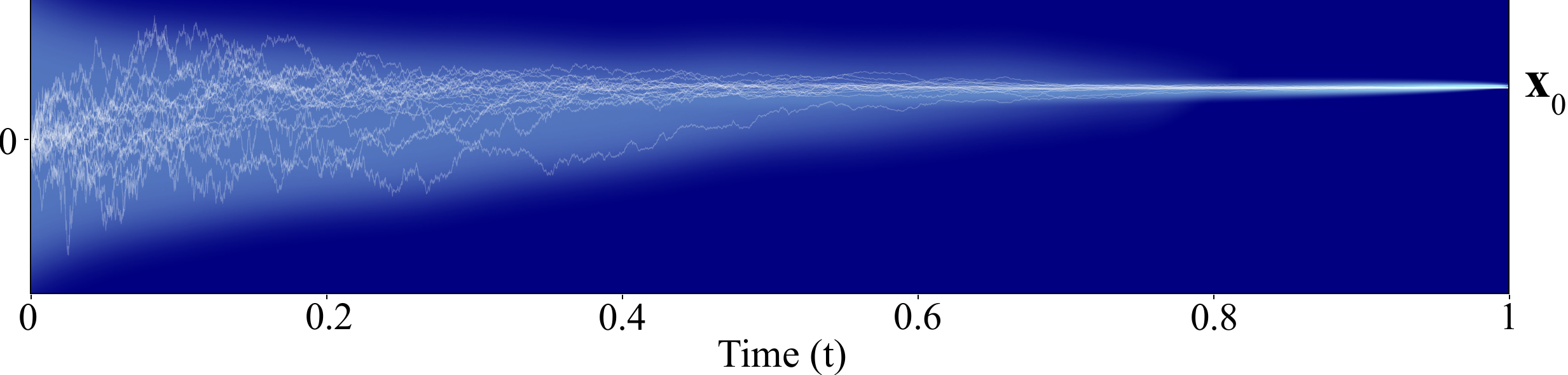}
    \caption{This figure illustrates optimization in the parameter space after $t$ iterations.}
    \label{fig:paramSearch}
\end{figure}

Figure \ref{fig:paramSearch} illustrates the optimal data distribution learned in the parameter space. The plot presents stochastic parameter trajectories for the input distribution mean (indicated by white lines) overlaid on a Bayesian flow distribution logarithmic heatmap.
\section{ELBO of ParamReL}
\label{app:der_ParamReL}
We derive the ELBO of ParamReL defined in Eq.~(\ref{eq:ParamReL-elbo}).
\begin{align}
\label{eq:ParamReL-elbo-full}
&\quad \log p(\mbx_0)\nonumber \\
& = \log \int_{\{\mbz_t\}_t} \int_{\{\mbx_t\}_t}  p\left({\mbx}_{0}, \{{\mbx}_t\}_{t}, \{\mbz_t\}_t\g\mbtheta_0, \alpha\right) \mathrm{d} \{\mbz_t\}_t \mathrm{d} \{\mbx_t\}_t \nonumber \\
& = \log \int_{\{\mbtheta_t\}_t}\int_{\{\mbz_t\}_t} \int_{\{\mbx_t\}_t} p(\{\mbtheta_t\}_t|-)p_\mathrm{O}(\mbx_0;\psi(\mbtheta_0, \mbz_0))\prod_{t=T}^1p(\mbz_t)\mathbb{E}_{p_\mathrm{O}(\mbx_t;\psi(\mbtheta_t, \mbz_t))}[p_\mathrm{S}(\mbx_{t-1}\g\mbx_t;\alpha_t)]\nonumber \\
& \qquad \qquad \mathrm{d} \{\mbz_t\}_t \mathrm{d} \{\mbx_t\}_t\mathrm{d} \{\mbtheta_t\}_t \nonumber \\
& = \log \int_{\{\mbz_t\}_t} \int_{\{\mbx_t\}_t} \int_{\{\mbtheta_t\}_t} p(\{\mbtheta_t\}_t|-)\frac{p_\mathrm{O}(\mbx_0;\psi(\mbtheta_0,\mbz_0))\prod_{t=T}^1p(\mbz_t)\mathbb{E}_{p_\mathrm{O}(\mbx_t;\psi(\mbtheta_t, \mbz_t))}[p_\mathrm{S}(\mbx_{t-1}\g\mbx_t;\alpha_t)]}{\prod_{t=1}^T p_\mathrm{S}(\mbx_{t-1}\g\mbx_t;\alpha_t)q_{\mbphi}(\mbz_t|\mbtheta_t,t)}\nonumber \\
& \qquad \qquad  \qquad \cdot \prod_{t=1}^T p_\mathrm{S}(\mbx_{t-1}\g\mbx_t;\alpha_t)q_{\mbphi}(\mbz_t|\mbtheta_t,t)\mathrm{d} \{\mbz_t\}_t \mathrm{d} \{\mbx_t\}_t \mathrm{d} \{\mbtheta_t\}_t \nonumber \\
& \ge \mathbb{E}_{\prod_{t=1}^T p_\mathrm{S}(\mbx_{t-1}\g\mbx_t;\alpha_t)q_{\mbphi}(\mbz_t|\mbtheta_t,t)p(\mbtheta_t|-)}\left[\log \frac{p_\mathrm{O}(\mbx_0;\psi(\mbtheta_0, \mbz_0))\prod_{t=T}^1p(\mbz_t)\mathbb{E}_{p_\mathrm{O}(\mbx_t;\psi(\mbtheta_t, \mbz_t))}[p_\mathrm{S}(\mbx_{t-1}\g\mbx_t;\alpha_t)]}{\prod_{t=1}^T p_\mathrm{S}(\mbx_{t-1}\g\mbx_t;\alpha_t)q_{\mbphi}(\mbz_t|\mbtheta_t,t)}\right]\nonumber \\
&= \sum_{t=1}^T \mathbb{E}_{p_F(\mbtheta_{t}|-)}\mathbb{E}_{q_{\mbphi}(\mbz_t)}\left\{\E_{p_\mathrm{S}\left(\mbx_{t-1} \g{\mbx_0};\alpha_{T:t}\right)}\left[\log\frac{p_\mathrm{S}\left(\mbx_{t-1} \g{\mbx_0};\alpha_{T:t}\right)}{p_{\mathrm{R}}\left(\mbx_{t-1} ;\psi(\mbtheta_t, \mbz_t), \alpha_t\right)}\right]\right.  \nonumber \\
 &\qquad \qquad \qquad \left.- \E_{q_{\mbphi}\left({\mbz}_t \g {\mbtheta_t}\right)}\left[\log\frac{q_{\mbphi}\left({\mbz}_t \g {\mbtheta_t}\right)}{p({\mbz}_t)}\right]\right\}+ \mathbb{E}_{q_{\mbphi}(\mbz_0,\mbtheta_0)}\left[{\ln p_\mathrm{O}(\mbx_0;\psi(\mbtheta_0, \mbz_0))}\right]\nonumber \\
&= -\sum_{t=1}^T \mathbb{E}_{p_F(\mbtheta_{t}|-)}\mathbb{E}_{q_{\mbphi}(\mbz_t)}\left\{{\KL{p_\mathrm{S}\left(\mbx_{t-1} \g{\mbx_0};\alpha_{T:t}\right)}{p_{\mathrm{R}}\left(\mbx_{t-1} ;\psi(\mbtheta_t, \mbz_t), \alpha_t\right)}}\right.  \nonumber \\
 &\qquad \qquad \qquad \left.- \KL{q_{\mbphi}\left({\mbz}_t \g {\mbtheta_t}\right)}{p({\mbz}_t)}\right\}+ \mathbb{E}_{q_{\mbphi}(\mbz_0,\mbtheta_0)}\left[{\ln p_\mathrm{O}(\mbx_0;\psi(\mbtheta_0, \mbz_0))}\right]:=\mathcal{L}_{\texttt{ParamReL}}
\end{align}



\section{Illustration of Network Architecture}
\label{app:enc_fram}

\subsection{Latent Encoding}
We define the encoder $q_{\mbphi}(\mbz_t|\mbtheta_t, t)$ with a latent encoding network parameterized by a U-Net, and 

the decoder as:
\begin{equation}
p_\mathrm{O}(\mbx_t;\psi(\mbtheta_t, \mbz_{t}))=\delta\left({\mbx}-\frac{\boldsymbol{\mu}}{\gamma(t)}-\sqrt{\frac{1-\gamma(t)}{\gamma(t)}} \epsilon_{\psi}({\mbtheta}_t, \mbz_{t})\right),
\end{equation}
with a noise prediction network $\epsilon_{\psi}({\mbtheta}_{t}, \mbz_{t})$ parameterized by a U-Net ~\cite{ronneberger2015u} same with ~\cite{ho2020denoising}.
Following the previous work ~\cite{dhariwal2021diffusion}, we employ an adaptive group normalization layer (AdaGN) to do conditional embedding, incorporating the input parameters and timestep into each residual block after a group normalization:
\begin{equation}
\operatorname{AGN}(\mbtheta_t, t)=(1+\mathbf{s}(t)) \cdot \operatorname{GroupNorm}(\mbtheta_t)+\mathbf{b}(t),
\end{equation}
where the $\mathbf{s}(\cdot)$ function outputs scaling factors based on the conditional input $t$, dynamically adjusting the normalized features to adapt to different contexts, and the $\mathbf{b}(\cdot) $ function provides bias terms derived from $t$, allowing shifting of the normalized features according to the given conditions. In decoders, the AGN layer will be as follows:
\begin{equation}
\operatorname{AGN}(\mathbf{\theta}_t, \mathbf{z}_t, t) = \left(1 + \mathbf{s}(t)\right) \cdot \operatorname{GroupNorm}\left(\left(1 + \mathbf{s}(\mathbf{z}_t)\right) \cdot \operatorname{GroupNorm}(\mathbf{\theta}_t) + \mathbf{b}(\mathbf{z}_t)\right) + \mathbf{b}(t).
\end{equation}

\subsection{Modified U-Net}
\begin{figure}[!htbp]
    \centering
    \includegraphics[width=1\columnwidth]{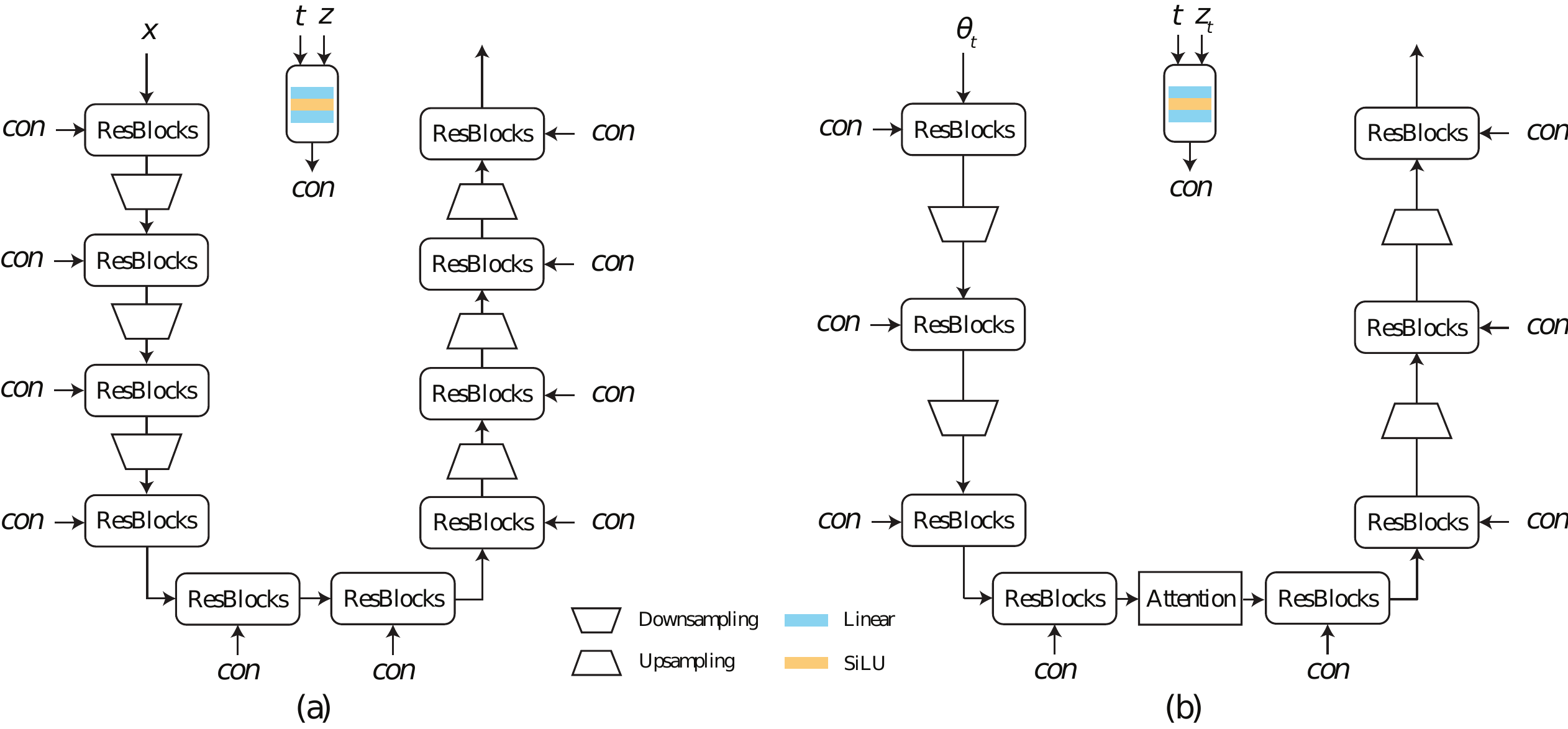}
    \caption{U-Net comparisons of infoDiffusion (a) and ours (b). We apply the Attention module in the bottleneck layer, shallower than the infoDiffusion's U-Net.  }
    \label{fig:unet}
\end{figure}

Similar to the diffusion-based representation learning model, we update the U-Net architecture based on Residual Blocks and Attention Modules. However, unlike previous approaches \cite{ho2020denoising,song2020denoising,preechakul2022diffusion,wang2023infodiffusion}, we use shallower layers in the upper and down modules while incorporating an additional attention mechanism in the bottleneck module to achieve significant representations. Figure \ref{fig:unet} illustrates the specific structural differences.

\subsection{Encoder}
\begin{figure}[!htbp]
    \centering
    \includegraphics[width=1\columnwidth]{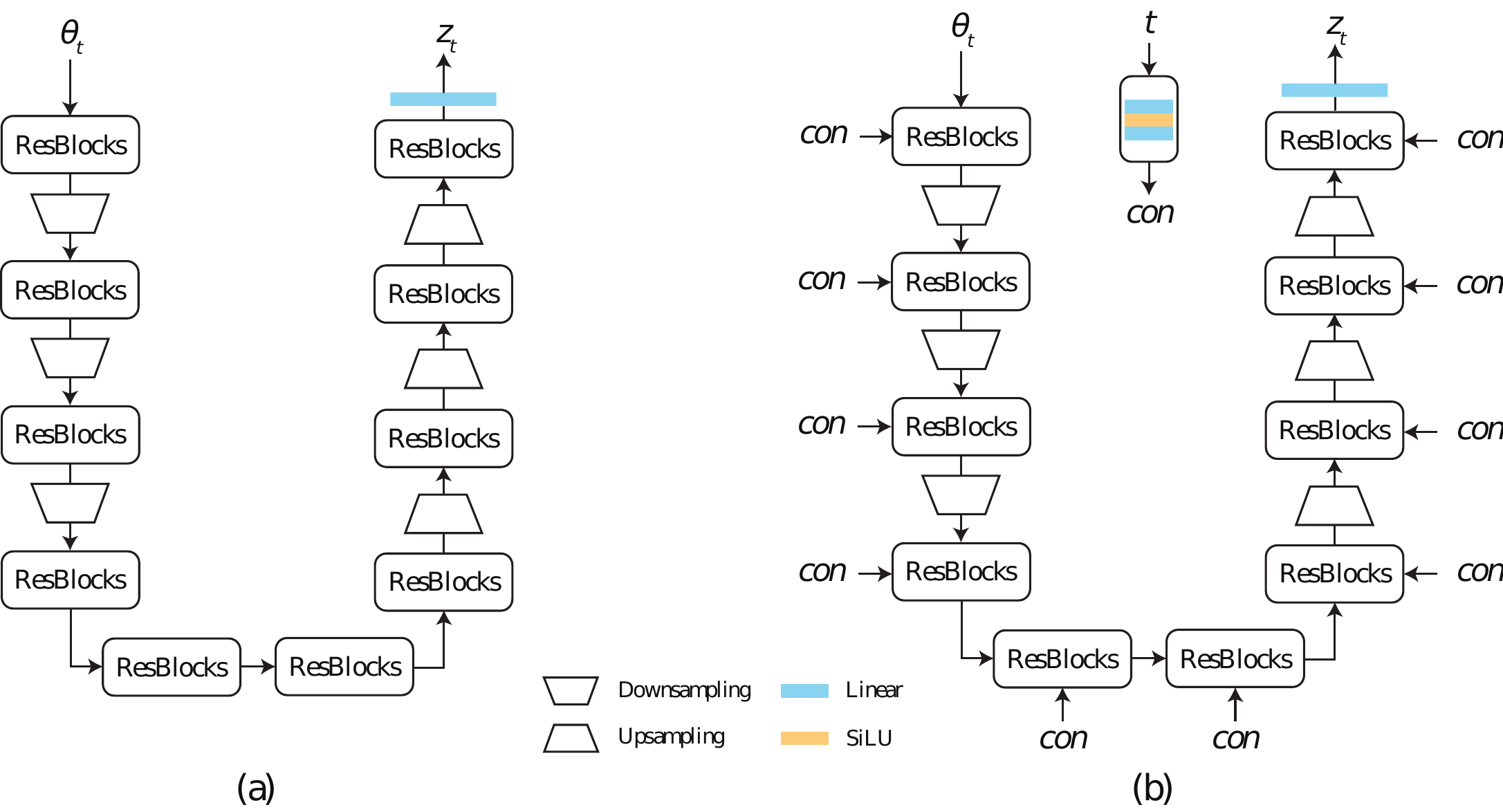}
    \caption{Two kinds of Encoders in ParamReL, including vanilla encoders (a) and progressive encoders (b) conditioned on time $t$. }
    \label{fig:encoder}
\end{figure}

We apply the same U-Net frameworks to encoders. The framework of encoders is depicted in Figure \ref{fig:encoder}.

The sizes of the model parameters for each configuration are provided in Table \ref{tab:para}.
\begin{table}[!htbp]
\centering
\label{tab:para}
\caption{Comparisons of parameters in training. We report the parameters of infoDiffusion and  ParamReL on the CelebA dataset with $(3 \times 64 \times 64)$ input size. }
\begin{tabular}{l|lll}
\toprule[1pt]
 &       & \textbf{Encoder} & \textbf{Decoder} \\
 \hline
 \textbf{CelebA} & \textbf{infoDiffusion} &   143.89\texttt{MB}  &   200.18\texttt{MB} \\
 $(3\times64\times64)$ & \textbf{ParamReL}  &  245.41\texttt{MB}   & 324.31\texttt{MB}   \\
\toprule[1pt]
\end{tabular}

\end{table}

\section{Hyperparameters for Training}
Table \ref{table:hyper} presents the hyperparameter settings for training ParamReL. Different bin values are provided for various continuous datasets. All models are trained for 50 epochs. ``Channel mult" denotes the channel shapes in each ResNet block within the U-Net architecture.

\label{app:hyer}
\begin{table}[!htbp]
\centering
\small
\caption{Hyperparameters for training Bayesian Flow Networks, U-Net architecture, training protocols, and devices.}
\label{table:hyper}
\begin{tabularx}{\textwidth}{Xcccccccccc} 
\toprule[1pt]
\multicolumn{2}{c}{} & \multicolumn{2}{c}{\textbf{Discrete}} & \multicolumn{3}{c}{\textbf{Continuous}} \\
\cmidrule(lr){3-4} \cmidrule(lr){5-7} 
 & & bFashionMNIST & bMNIST  & CelebA & 3DShapes & CIFAR10 \\
\midrule
\multirow{1}{*}{\textbf{BFN}} & bins & -&- &256 &256 & 16 \\
\midrule
\multirow{4}{*}{\textbf{Exps}} & Latent Dim &32& 32& 64& 32 & 32\\
 & Batch Size & 128&128 & 64 &64 & 128\\
 & Learning Rate & 1e-4& 1e-4 &2e-4 &2e-4 &1e-4 \\
 & DataShape  & (28, 28, 1) & (28, 28, 1) & (64,64,3) & (64,64,3) & (32,32,3) \\
\midrule
\multirow{2}{*}{\textbf{Models}} & Channel Mult & [1,2,4]& [1,2,4]& [1,2,2,2] &[1,2,2,2] & [1,2,4]\\
 & Num. Channels & 32&32 & 64&64  &32 \\
 \hline
 
 \textbf{Device} & GPU & H100 &H100 & H100& H100& H100 \\
\toprule[1pt]
\end{tabularx}
\end{table}

\section{Latent Interpolation}
\label{app:inter}
The latent space interpolation can be described as follows. Firstly, we noise source images to generate latent pairs by sender distribution, $<\mbx^{1}_{1}, \mbx^{2}_{1}>$, where $\mbx^{1}_{1} \sim q(\cdot \mid \mbx^{1}_{N}) $ and $\mbx^{2}_{1} \sim q(\cdot \mid \mbx^{2}_{N})$. Then, we implement two methods from~\cite{shoemake1985animating} to generate four interpolated latent pairs $\bar{\mbx}_{1:4}$, i.e., linear interpolation, and spherical interpolation:
\begin{equation}
\begin{aligned}
\bar{\mbx}_i & =(1-\lambda) \mbx_{0}^{1}+\lambda \mbx_{0}^{2}, \\
\bar{\mbx}_i & =\frac{\sin ((1-\alpha) \theta)}{\sin (\theta)} \mbx_{0}^{1}+\frac{\sin (\alpha \theta)}{\sin (\theta)} \mbx_{0}^{1}, \\
\end{aligned}
\end{equation}
where $\lambda$ is the scale coefficient, $\alpha \in [0,1]$ denotes the interpolation steps, and $\theta=\arccos \left(\frac{\left(\mbx_{0}^{1}\right)^{\top} \mbx_{0}^{2}}{\left\|\mbx_{0}^{1}\right\|\left\|\mbx_{0}^{2}\right\|}\right)$ is the angle between $\mbx_{0}^{1}$ and $\mbx_{0}^{2}$.

\begin{figure}[H]
    \centering
    \includegraphics[width=0.7\columnwidth]{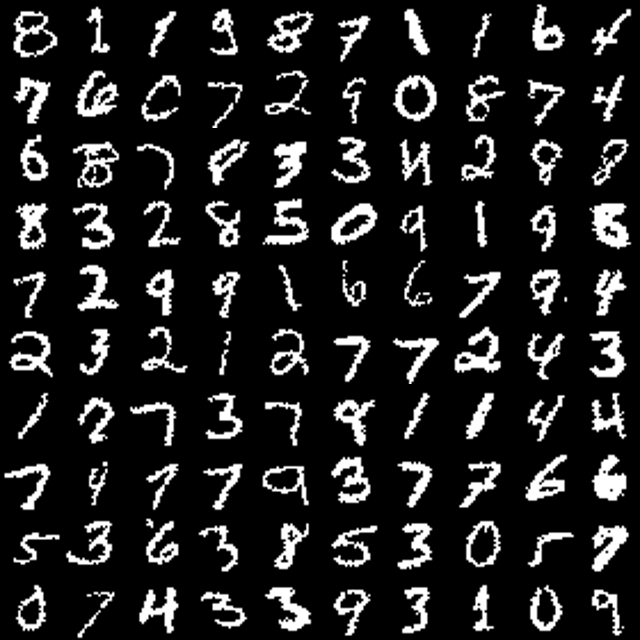}
    \caption{Samples reconstructed from our trained ParamReL on dataset Binary-MNIST.}
    \label{fig:recon-bfn-mnist}
\end{figure}

\begin{figure}[H]
    \centering
    \includegraphics[width=1\columnwidth]{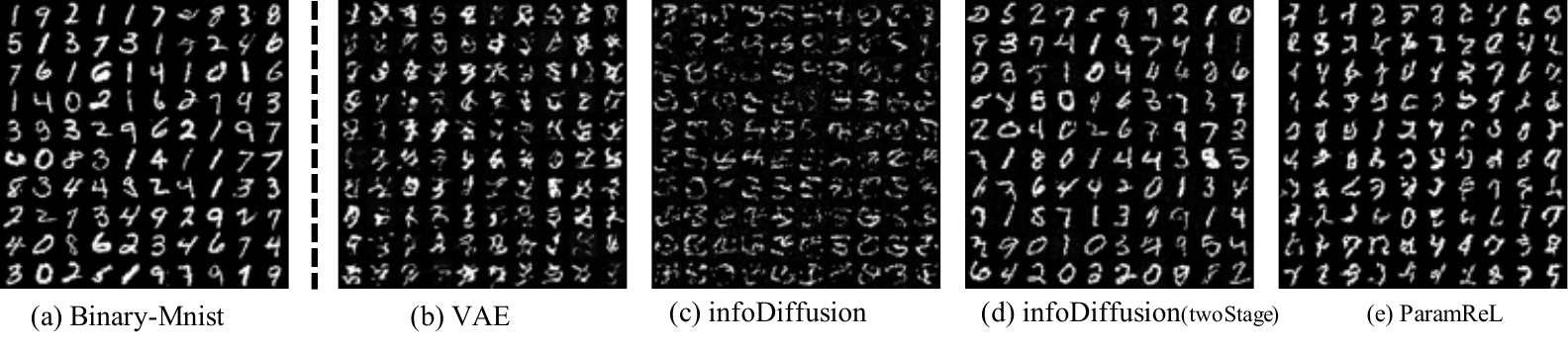}
    \caption{Samples generated from our trained ParamReL.}
    \label{fig:gen-bfn}
\end{figure}
\begin{figure}[H]
    \centering
    \includegraphics[width=0.8\columnwidth]{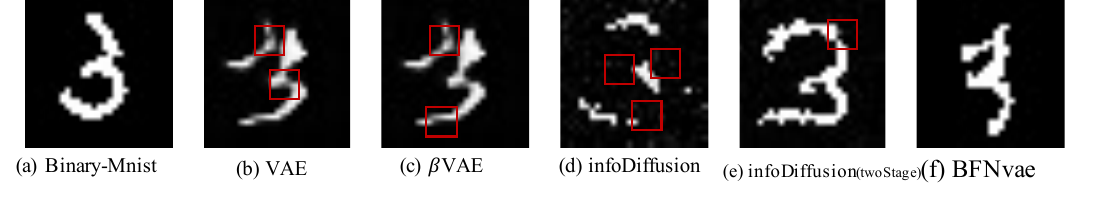}
    \caption{Details of samples generated from our trained ParamReL.}
    \label{fig:gen2-bfn}
\end{figure}

\section{Generation Quality}
\label{app:gen}
We illustrate the unconditional generation quality in Figure \ref{fig:gen-bfn} and Figure \ref{fig:gen2-bfn} on bMNIST. Images sampled from VAE-based model are blurry, as shown in Figure \ref{fig:gen-bfn} (b). We implement two sampling strategies in the Diffusion-based model ~\cite{wang2023infodiffusion}, and both can only sample grey-scale images. Figure \ref{fig:gen-bfn} (c) is sampled from the DDIM sampler, and Figure \ref{fig:gen-bfn} (d) is sampled from a two-phased sampling procedure: form timesteps $T$ to $T/2$,  denoise and sample using a pre-trained vanilla denoising diffusion model. For timesteps ranging from $T/2$ to $0$, proceed with sampling utilizing the InfoDiffusion model. Figure \ref{fig:gen-bfn} (e) is images generated from our ParamReLc model. Refer to Figure \ref{fig:gen2-bfn} for more detailed information per image sampled from corresponding models. We can conclude that ParamReL can be sampled from the discrete distribution where the image value is binarized. 
\begin{figure}[H]
    \centering
    \includegraphics[width=0.8\columnwidth]{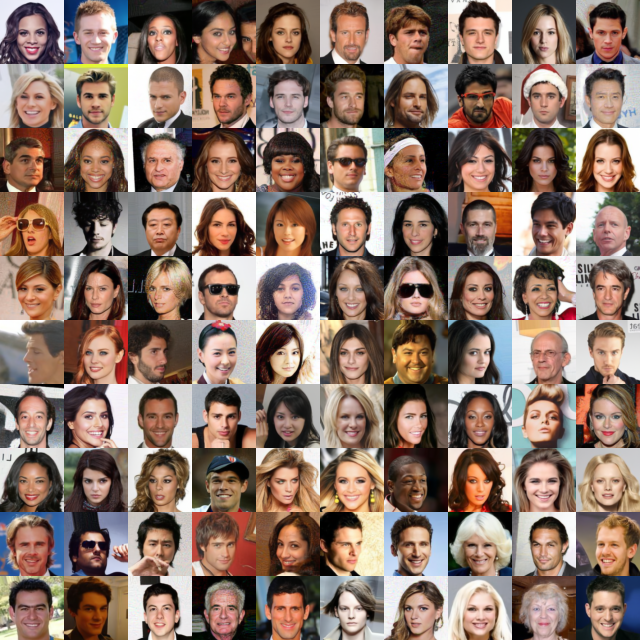}
    \caption{Samples reconstructed from our trained ParamReL on CelebA.}
    \label{fig:recon-bfn-celeba}
\end{figure}

\section{Reconstruction Quality}
\label{app:recon}
\begin{figure}[H]
    \centering
    \includegraphics[width=0.8\columnwidth]{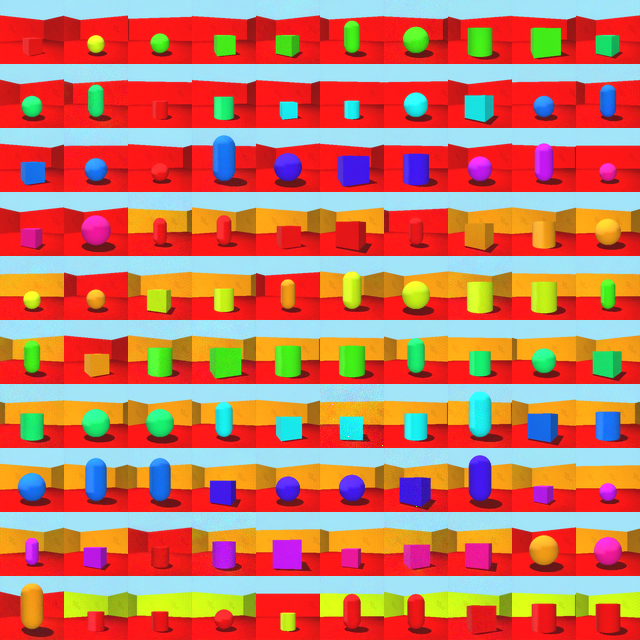}
    \caption{Samples reconstructed from our trained ParamReL on 3DShapes.}
    \label{fig:recon-bfn-3dshapes}
\end{figure}

The qualitative reconstruction results are shown in Figure \ref{fig:recon-bfn-celeba} for CelebA, Figure \ref{fig:recon-bfn-3dshapes} for 3DShapes  and Figure \ref{fig:recon-bfn-3dshapes} for CIFAR10.

\begin{figure}[H]
    \centering
    \includegraphics[width=0.8\columnwidth]{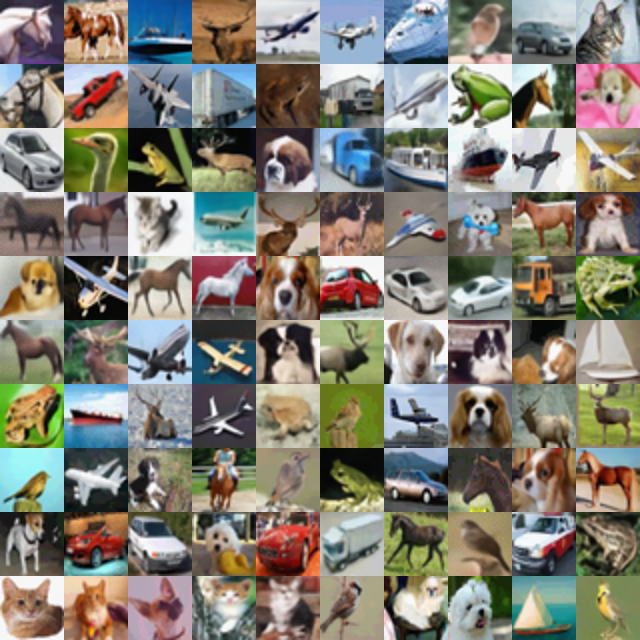}
    \caption{Samples reconstructed from our trained ParamReL on CIFAR10.}
    \label{fig:recon-bfn-cifar10}
\end{figure}

\end{document}